\documentclass[conference]{IEEEtran}

\IEEEoverridecommandlockouts

\usepackage{cite}
\usepackage{amsmath,amssymb,amsfonts}
\usepackage{graphicx}
\usepackage{textcomp}
\usepackage{xcolor}

\usepackage{latexsym}
\usepackage{amsthm}
\usepackage{booktabs}
\usepackage{enumitem}
\usepackage{bm}
\usepackage{algorithm}
\usepackage{algorithmicx}
\usepackage{algpseudocode}
\usepackage{subfigure}
\usepackage{multirow}
\usepackage{xspace}
\usepackage{url}

\newcommand{\sysname}{FedMCP\xspace}

\begin{document}

\title{FedMCP: Parameter-Efficient Federated Learning with Model-Contrastive Personalization}

\author{\IEEEauthorblockN{Qianyi Zhao$^1$, Chen Qu$^2$, Cen Chen$^{1,*}$\thanks{$^*$Corresponding author.}, Mingyuan Fan$^1$, Yanhao Wang$^1$}
\IEEEauthorblockA{
$^1$\textit{School of Data Science and Engineering, East China Normal University, Shanghai, China}\\
$^2$\textit{Manning College of Information \& Computer Sciences, University of Massachusetts Amherst, Amherst, MA, USA}\\
\texttt{51255903037@stu.ecnu.edu.cn, mail@cqu.org, cenchen@dase.ecnu.edu.cn,}\\
\texttt{fmy2660966@gmail.com, yhwang@dase.ecnu.edu.cn}}
}

\maketitle

\begin{abstract}
With increasing concerns and regulations on data privacy, fine-tuning pretrained language models (PLMs) in federated learning (FL) has become a common paradigm for NLP tasks.
Despite being extensively studied, the existing methods for this problem still face two primary challenges.
First, the huge number of parameters in large-scale PLMs leads to excessive communication and computational overhead.
Second, the heterogeneity of data and tasks across clients poses a significant obstacle to achieving the desired fine-tuning performance.
To address the above problems, we propose \sysname, a novel parameter-efficient fine-tuning method with model-contrastive personalization for FL.
Specifically, \sysname adds two lightweight adapter modules, i.e., the \emph{global adapter} and the \emph{private adapter}, to the frozen PLMs within clients.
In a communication round, each client sends only the global adapter to the server for federated aggregation.
Furthermore, \sysname introduces a model-contrastive regularization term between the two adapters.
This, on the one hand, encourages the global adapter to assimilate universal knowledge and, on the other hand, the private adapter to capture client-specific knowledge.
By leveraging both adapters, \sysname can effectively provide fine-tuned personalized models tailored to individual clients.
Extensive experiments on highly heterogeneous cross-task, cross-silo datasets show that \sysname achieves substantial performance improvements over state-of-the-art FL fine-tuning approaches for PLMs.
\end{abstract}


\section{Introduction}

Pretrained language models (PLMs) have recently gained considerable attention for their wide applications in various natural language processing (NLP) tasks.
Fine-tuning PLMs on specific datasets is often essential to ensure good performance for downstream tasks.
However, due to increasing concerns and regulations about \emph{data privacy}, the datasets are usually distributed among multiple entities, forming private data silos across different clients \cite{quNaturalLanguageUnderstanding2021}.
When fine-tuning PLMs, clients are not allowed to share their private datasets with the central server or other clients.
For example, Rieke et al.~\cite{rieke2020future} note that data silos are common in the healthcare domain, where patient information is critical to training diagnostic or treatment recommendation models but is isolated among multiple healthcare institutions.
To address the above issue, federated learning (FL) \cite{konevcny2016federated, pmlr-v54-mcmahan17a} has emerged as a promising solution by allowing different clients to collaboratively train PLMs without sharing local private data \cite{lin2021fednlp}.

FL encounters several obstacles in the context of PLM fine-tuning.
One significant challenge is the limited communication bandwidth and computational resources on client devices.
In particular, FL involves frequent model exchanges between the central server and clients during the training process.
Due to the huge number of parameters in large PLMs, these exchanges can lead to a high communication overhead.
Furthermore, the tight computational resources of clients make fine-tuning all parameters in the PLM unaffordable \cite{zhang-etal-2023-fedpetuning}.
This poses a barrier to the deployment of large PLMs, such as BERT \cite{devlin2018bert}, GPT \cite{radford2019glue}, and T5 \cite{raffel2020exploring} in federated settings \cite{wu2022communication}.

Another challenge lies in the data and task heterogeneity among clients, known as the non-independent-and-identically distributed (non-IID) problem.
Typical FL methods, such as FedAvg \cite{pmlr-v54-mcmahan17a}, train a unified global model for all participants.
However, due to the variety of data and tasks across clients, the global model may be suboptimal for each client \cite{huang2021personalized}.
The common strategy to mitigate the non-IID problem is \textit{model personalization} \cite{tan2022towards}, which tailors the global model to suit the specific needs and data characteristics of individual clients.
Existing model personalization methods mainly address non-IID scenarios with different data and label distributions among clients. 
However, real-world NLP systems encompass more complex non-IID scenarios \cite{ye2023heterochallenges}, where different clients hold textual data in different domains, such as question answering, social posts, emails, etc., each focusing on its specific tasks.
Heterogeneity of this type presents more serious challenges to FL but is under-explored in the literature.

To address the above challenges, in this paper, we propose \sysname, a novel \underline{Fed}erated learning method with \underline{M}odel-\underline{C}ontrastive \underline{P}ersonalization that aims to fine-tune PLMs in a parameter-efficient manner to reduce communication and computational costs while mitigating the heterogeneity of data and tasks for natural language understanding (NLU) in cross-silo settings.
In general, \sysname adopts a common paradigm for personalized FL \cite{tan2022towards}, which collectively trains a global model to learn universal knowledge that is independent of data distributions and specific tasks, and then personalizes the global model via local adaptation to capture data- and task-specific knowledge within each client.

\begin{figure}[t]
  \centering
  \includegraphics[width=0.45\textwidth]{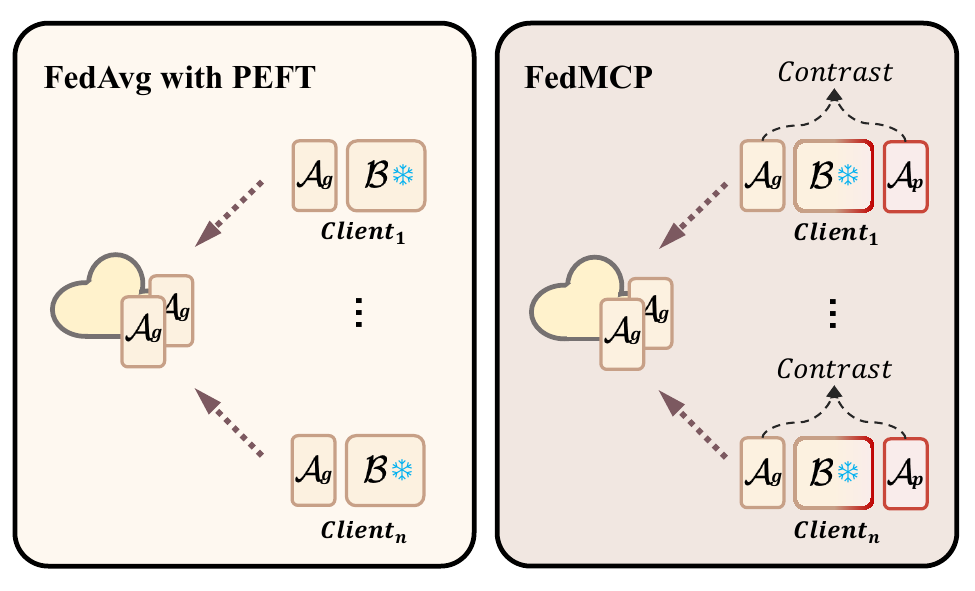}
  \caption{Comparison of FedAvg with PEFT and \sysname, where $\mathcal{A}$ and $\mathcal{B}$ refer to the adapter and backbone modules, respectively, and the snowflake icon indicates that the backbone is frozen, with only the adapters trainable.}
  \label{fig1}
\end{figure}

Specifically, \sysname incorporates two adapter modules \cite{houlsbyadapter2019}, i.e., \emph{global} and \emph{private adapters}, into the PLM backbone for personalization.
In each communication round, only the global adapter participates in the federated aggregation process to facilitate collaboration and knowledge sharing among different clients.
Meanwhile, the private adapter remains local to learn client-specific knowledge. 
Moreover, we introduce a novel model-contrastive personalization loss tailored to the parameter-efficient fine-tuning (PEFT) method for FL.
This loss function leverages central kernel alignment (CKA) \cite{kornblith2019similarity} to measure similarities between adapter modules.
By minimizing the distance between the global adapter of each client and the average global adapter, while maximizing the distance between the global and private adapters of each client, \sysname achieves a good trade-off between model generalization and personalization in the sense that the global adapter learns universal knowledge, whereas the private adapter captures the knowledge specific to each client.
Fig.~\ref{fig1} illustrates \sysname compared to the widely used FedAvg (with PEFT). When fine-tuning the PLM in a federated setting, FedAvg with PEFT keeps the backbone fixed and trains the adapter module $\mathcal{A}_g$; \sysname introduces an additional trainable adapter module $\mathcal{A}_p$ that is not involved in the federated aggregation and employs a model contrastive learning method on the two adapters to train personalized models.

Finally, we conduct extensive experiments to evaluate the efficacy and efficiency of \sysname.
We use six datasets from the GLUE benchmark \cite{radford2019glue} to simulate the cross-task, cross-silo scenario, where each client holds a specific type of textual data and focuses on a distinct NLU task.
The experimental results demonstrate that \sysname outperforms several state-of-the-art personalized FL methods by approximately 1.5\% in terms of the average client accuracy.
Moreover, \sysname with PEFT achieves a performance comparable to fine-tuning the entire PLM while significantly reducing communication and computational costs.

Our main contributions are summarized as follows:
\begin{itemize}
  \item We propose \sysname, a novel parameter-efficient personalized FL method that utilizes the global and private adapters to mitigate the heterogeneity of data and tasks for NLU in the PLM fine-tuning.
  \item We present a model-contrastive personalization loss for \sysname to achieve a good trade-off between generalization to universal knowledge and personalization to client-specific knowledge.
  \item We conduct comprehensive experiments on the composed dataset to verify the superior performance of \sysname compared to state-of-the-art personalized FL methods. 
\end{itemize}
\section{Related Work}

\subsection{(Personalized) Federated Learning}

Seminal FL training schemas such as FedAvg \cite{pmlr-v54-mcmahan17a} aggregate local models into the global model via averaging.
However, they suffer from unstable convergence and performance degradation in non-IID settings.
Therefore, several methods, such as FedProx \cite{li2020federated} and MOON \cite{li2021moon}, were proposed to address the non-IID issue by constraining local updates with the global model.
Particularly, FedProx constrains local updates using $l_2$-distances; MOON leverages the heterogeneity in the representations learned by individual clients compared to the global model for local update correction.
However, they still provide a single global model and may not adequately meet the requirements of different clients in non-IID settings.
This leads to the emergence of personalized FL methods.

The existing personalized FL methods can be broadly classified into four categories based on the techniques they used, namely \emph{distillation}, \emph{regularization}, \emph{adaptive collaboration}, and \emph{parameter decoupling}.
FedMD \cite{li2019fedmd} and FedDF \cite{sattler2020FEDdf} utilized knowledge distillation for model personalization.
pFedMe \cite{t2020personalized} and Ditto \cite{li2021ditto} introduced regularization terms based on meta-learning and multi-task learning, respectively, to prevent client models from overfitting to local data by comparing them to the global model.
MOCHA \cite{huang2021personalized} and FedAMP \cite{zhang2020fedfomo} proposed adaptive schemes to encourage clients with similar data distributions to collaborate more.
The methods in \cite{arivazhagan2019fedper,oh2021fedbabu,collins2021FEDREP} decoupled the network by retaining the parameters in personalized layers locally for individual clients while sharing only the global parameters for aggregation.
In particular, FedPer \cite{arivazhagan2019fedper} divided a deep feedforward neural network into shared base layers and personalized layers; FedBABU \cite{oh2021fedbabu} and FedRep \cite{collins2021FEDREP} adopted another scheme that divides the neural network into a shared body to learn global feature representation across clients and unique local heads for personalized classification in each client.
In this paper, the \sysname method extends the high-level idea of parameter decoupling through contrastive personalization and combines it with PEFT for PLMs.

\subsection{Federated Learning for NLP}

FL has also been widely used for NLP tasks, including news recommendation \cite{yi2021efficient}, question answering \cite{chen2021fedmatch,dong2022FEDATC}, and text summarization \cite{pan2023FedSUMM}.
In these applications, PLMs are shown to be effective in generating text representations that capture useful knowledge for downstream tasks.
For example, FedMatch \cite{chen2021fedmatch} introduced a backbone-patch architecture, where the shared backbone learns common knowledge and the private patch holds information specific to each client.
However, exchanging all PLM parameters during FL training requires substantial computational and communication resources.

Due to resource limitations, FL methods that can effectively train PLMs with high computational and communication efficiencies have recently attracted much attention.
Passban et al.~\cite{passban2022training} first introduced domain adapters to neural machine translation (NMT) models in federated settings.
Fed-MNMT \cite{liu2023communication} considered fine-tuning PLMs with adapters for multilingual NMT, alleviating data discrepancy effects through clustering strategies.
However, the above methods are limited to NMT but do not consider any other NLP tasks.
FedPETuning \cite{zhang-etal-2023-fedpetuning} investigated the performance of PEFT methods for PLMs in FL settings.
C2A \cite{kim-etal-2023-C2A} further proposed a hypernetwork-based framework to generate client-customized adapters to reduce client drifts in PEFT approaches.
However, they only consider the heterogeneity of data and label distributions among clients, which show subpar performance in the cross-silo scenario with distinct client-level tasks.

\begin{figure*}[t]
  \centering
  \includegraphics[width=\linewidth]{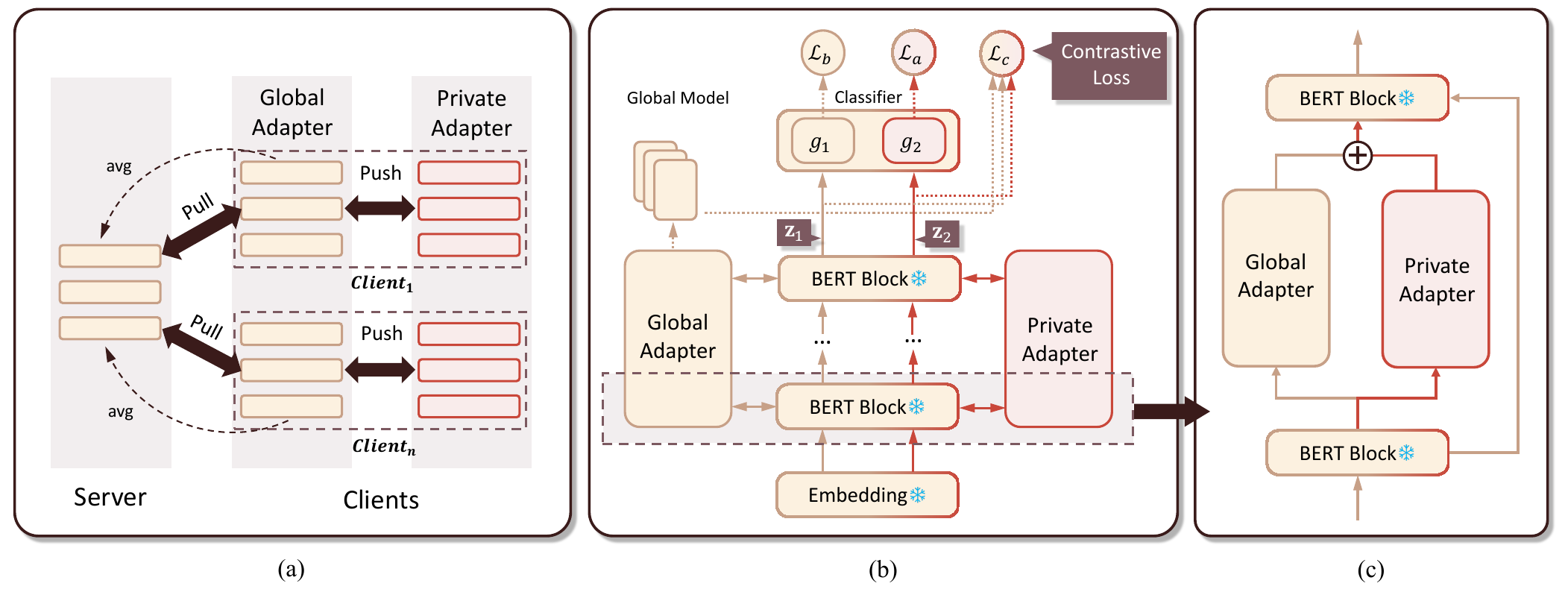}
  \caption{Overview of the \sysname method. (a) Federated model-contrastive personalization workflow; (b) Overall model structure; (c) Detailed structure of the two adapters and BERT blocks.}
  \label{fig:overview}
\end{figure*}

\section{Preliminaries}

In this section, we introduce the background of FL and PEFT for NLU tasks.

\smallskip
\noindent\textbf{Federated Learning for NLU.}
This paper focuses on NLU in the federated setting, specifically on \emph{supervised text classification} tasks, following previous studies \cite{luo2021ccvr, xu2023fedpac}.
The model is decomposed into a \emph{text encoder} and \emph{task-specific classifiers}.
Suppose that there are $m$ clients with the $i$-th client having a data distribution $P_{XY}^i$ on $\mathcal{X} \times \mathcal{Y}$, where $\mathcal{X}$ and $\mathcal{Y}$ are the input space and the label space, respectively.
Given a sample $(\mathbf{x}, \mathbf{y})$, the text encoder $f_{\theta} : \mathcal{X} \to \mathcal{Z}$ (parameterized by $\theta$) maps the input $\mathbf{x}$ to a feature vector $\mathbf{z} = f_{\theta}(\mathbf{x}) \in \mathbb{R}^d$ in the feature space $\mathcal{Z}$.
Subsequently, the classifier $g_{\phi} : \mathcal{Z} \to \mathcal{Y}$ (parameterized by $\phi$) maps the feature $\mathbf{z}$ to predict the label $g_{\phi}(\mathbf{z}) \in \mathcal{Y}$.
The parameters of the whole model are represented by $w = (\theta, \phi)$.
In the $t$-th round of FL, the server broadcasts the model parameters $w^{t-1}$ after the $(t-1)$-th round to all clients.
Then, the $i$-th client locally optimizes the following objective to obtain $w^{i,t}$:
\begin{equation}
\label{eq: local objective.}
    \min_{w^{i,t}} \mathbb{E}_{(\mathbf{x},\mathbf{y}) \sim P^i_{XY}}[\mathcal{L}(w^{i, t}; w^{i, t-1}, \mathbf{x}, \mathbf{y})],
\end{equation}
where $\mathcal{L}$ denotes the loss function. 
After local training, the server collects the updated models from participating clients and aggregates them into the global model.
The above process is performed iteratively until convergence.

\smallskip
\noindent\textbf{PEFT with Adapters.}
Introducing additional parameters with adapters \cite{houlsbyadapter2019} is a common paradigm to fine-tune PLMs in a parameter-efficient manner.
Taking Transformer-based PLMs \cite{devlin2018bert} as an example, an adapter is added after the attention and feedforward network layers in the form of a fully connected network.
This method demonstrates high parameter efficiency by updating only a small subset of parameters during fine-tuning while achieving performance comparable to fully fine-tuning all parameters.
For a hidden layer output $\mathbf{h}$, the down-projection layer $\mathbf{W}^{\text{down}}$ of the adapter layer projects $\mathbf{h}$ to a space with a lower dimension $r$.
Subsequently, a non-linear activation function such as GeLU \cite{hendrycks2016gaussian} is used to map the vector back to the same dimension as that of $\mathbf{h}$ through an up-projection $\mathbf{W}^{\text{up}}$, and the computation process of the adapter can be represented as
\begin{equation}
    \mathbf{h} \gets \mathbf{h} + \mathrm{GeLU}(\mathbf{h}\mathbf{W}^{\text{down}})\mathbf{W}^{\text{up}}.
\label{eq: adapter}
\end{equation}
Our basic idea in this work is also to incorporate adapters into the model and employ effective tailor strategies to make them learn knowledge specific to each client for personalization.

\section{Our Method}

In this section, we describe the proposed personalized FL method \sysname in detail.
We start with an overview (Section \ref{sec: overview}), followed by a description of the model architecture and the design of the two adapters (Section \ref{sec: Model Architecture}). 
Then, we show how client personalization is achieved by employing the model-contrastive method (Sections \ref{sec: global adapter} and~\ref{sec: Contrastive Personalization}).
Finally, we provide the complete algorithmic procedure and the optimization objective for each client (Section \ref{sec: local training and aggregation}).

\subsection{Overview}
\label{sec: overview}

In this section, we provide an overview of the model and key ideas of our model-contrastive personalization approach.
Fig.~\ref{fig:overview}(b) illustrates the model structure and the components of the loss function.
The model consists of a backbone and two integrated adapter modules.
The client-side loss function during training comprises three components: the cross-entropy loss of the full model $\mathcal{L}_a$, the cross-entropy loss of the backbone with the global adapter $\mathcal{L}_b$, and the contrastive loss $\mathcal{L}_c$ between two adapters.
The aforementioned losses are introduced with the following two key considerations:
\begin{itemize}
    \item \textbf{Distinguishing local and global knowledge:} For the client-side model, the objective is to effectively distinguish local specific and global shared knowledge. This distinction primarily stems from the model-contrastive method we use.
    \item \textbf{Enhancing the representation power of the shared global adapter:} For the shared global adapter, another objective is to improve its learning capability, allowing it to acquire generic knowledge beneficial to all clients.
\end{itemize}
The interplay of these three losses enables the local model to capture both considerations.
In the following sections, we will detail each component of the loss function.

\subsection{Model Architecture}
\label{sec: Model Architecture}

As depicted in Fig.~\ref{fig:overview}(b), the model comprises a backbone and two additional adapter modules that are integrated into the backbone.
Fig.~\ref{fig:overview}(c) illustrates how the two adapters are organized within each two BERT blocks.
The two adapters are inserted into the same position of the backbone, and their outputs are averaged to serve as input for the next layer.
 
Following Eq.~(\ref{eq: adapter}), the global adapter and the private adapter are denoted by $\mathbf{W}_g$ and $\mathbf{W}_p$, respectively.
The output $\mathbf{h}$ of the hidden layer after passing through the two adapters is:
\begin{equation*}
    \mathbf{h} \gets \mathbf{h} + \frac{1}{2} \mathrm{GeLU}(\mathbf{h}\mathbf{W}_{g}^{\text{down}})\mathbf{W}_{g}^{\text{up}} + \frac{1}{2} \mathrm{GeLU}(\mathbf{h}\mathbf{W}_{p}^{\text{down}})\mathbf{W}_{p}^{\text{up}}.
\end{equation*}
For given input $\mathbf{x}$, the model undergoes two forward propagations: one through the full model incorporating the two adapters (\textcolor[RGB]{201,71,58}{\textbf{red lines}} in Fig.~\ref{fig:overview}), and the other through the backbone with only the global adapter (\textcolor[RGB]{199,160,135}{\textbf{brown lines}} in Fig.~\ref{fig:overview}).
We denote $\mathbf{z}_1$ and $\mathbf{z}_2$ as the representations generated by the full model and the backbone with the global adapter, respectively. After encoding, the sequence representations are fed into distinct multi-layer perceptron (MLP) classifiers $g_1$ and $g_2$, respectively, to obtain classification results.

The cross-entropy loss of the local full model $\mathcal{L}_a$ with the two adapters is formulated as:
\begin{equation}
    \mathcal{L}_a = \ell((\theta _g, \theta _p, \phi_a);(\mathbf{x}, \mathbf{y})),
\end{equation}
where $\ell$ is the cross-entropy loss, $\theta _g$ is the parameters of the global adapter, $\theta _p$ is the parameters of the private adapter, and $\phi_a$ is the parameters of the classifier in the local model.

\subsection{Global Adapter Learning}
\label{sec: global adapter}

One of the focuses of our method is to enable the global adapter to adapt to each client's downstream tasks, even without the help of private adapters.
Unlike typical model training that calculates the overall loss, we specifically compute the cross-entropy loss from predictions processed only through the backbone and the global adapter.
This strategy ensures that the global adapter is precisely adapted to the diverse client requirements.
The cross-entropy loss based on the global adapter is used for regularization, enhancing the ability of the global adapter to acquire client-independent universal knowledge.
For an input $(\mathbf{x}, \mathbf{y})$, the definition of the backbone with the loss of the global adapter is:
\begin{equation}
    \mathcal{L}_b = \ell((\theta _g, \phi_b);(\mathbf{x}, \mathbf{y})),
\end{equation}
where $\phi_b$ denotes the parameters of the classifier for the backbone with the global adapter.

\subsection{Model-Contrastive Personalization}
\label{sec: Contrastive Personalization}

\noindent\textbf{Background on Model-Contrastive Personalization.}
First introduced by Li et al.~\cite{li2021moon}, the MOON method focuses on model-level contrast to reduce discrepancies between local and global models in FL, with the objective of mitigating model drift in non-IID scenarios.
However, training a single, averaged global model lacks personalization and thus impairs the performance of individual clients with heterogeneous data distributions and specific tasks.

To achieve personalization within the PEFT framework, beyond the global module's aggregation, client-specific customization is also crucial.
Therefore, we integrate two tunable adapter modules into frozen PLMs.
Fig.~\ref{fig:overview}(a) shows the model-contrastive workflow with the two adapters.
The input of the model-contrastive workflow is the representations generated with the backbone enhanced with different adapters, that is, the (local) global adapter $\mathbf{X} \in \mathbb{R} ^{n \times h}$, the (local) private adapter $\mathbf{Y} \in \mathbb{R} ^{n \times h}$, and the shared average global adapter $\mathbf{Z} \in \mathbb{R} ^{n \times h}$, where $n$ is the batch size and $h$ is the hidden layer size of the model.
Specifically, these representations are obtained by applying an average pooling to the token representations from the encoder's last layer with the corresponding adapter.

The model-contrastive personalization process contains two loss components. 
The first component minimizes the similarity between the private adapter $\mathbf{X}$ and the global adapter $\mathbf{Y}$ of the client to differentiate the knowledge they acquire.
The second component maximizes the similarity between the global adapter $\mathbf{Y}$ of the client and the averaged global adapter $\mathbf{Z}$ to reduce model drift during federated aggregation.
This ensures that the global adapter learns client-agnostic knowledge while the private adapter gains client-specific knowledge.
By combining both components, the contrastive $\mathcal{L}_c$ loss during the training procedure is expressed as:
\begin{equation}
    \mathcal{L}_c = \mathrm{Sim}(\mathbf{X},\mathbf{Y}) - \mathrm{Sim}(\mathbf{X},\mathbf{Z}),
    \label{eq: contrastive loss}
\end{equation}
where $\mathrm{Sim}(\cdot, \cdot)$ can be any similarity metric applicable to vector representations.

\begin{algorithm}[t] 
    \caption{\sysname}
    \label{alg:alg1}
    {\bf Input:} Communication round $T$; number of local epochs $E$; learning rate $\eta$; and number of clients $m$.\\
    {\bf Server Executes:}\\
    \vspace{-0.5cm}
    \begin{algorithmic}[1]
        \For{each round $t=1$ to $T$} 
            \For{each client $i$ {\bf in parallel}}
                \State Send the average global adapter $\theta_{g}^{t}$ to client $i$;
                \State $\theta_{g}^{i, t} \gets$ \textbf{LocalUpdate}($i$, $\theta_{g}^{t}$, $\theta_{p}^{t}$);
            \EndFor
            \State Compute $\theta_{g}^{t+1} = \frac{1}{m} \sum^{m}_{i=1} \theta_{g}^{i,t}$;
        \EndFor
    \end{algorithmic}
  {\bf LocalUpdate}($i$, $\theta_{g}^{t}$, $\theta_{p}^{t}$)\textbf{:}\\
  \vspace{-0.5cm}
  \begin{algorithmic}[1]
        \For{each local epoch $e=1$ to $E$}
            \State Receive the average global adapter $\theta_{g}^{t}$;
            \State Obtain the private adapter $\theta_{p}^{t}$;
            \State Compute $\mathcal{L}$ by Eq.~(\ref{eq: local overall loss});
            \State Update $\theta_{g}^{t}$, $\theta_{p}^{t}$ by Eq.~(\ref{eq: local update});
        \EndFor
        \State Set $\theta_{p}^{t + 1} \gets \theta_{p}^{t}$;
        \State \Return{$\theta_{g}^{i, t} \gets \theta_{g}^{t}$ to the server};
    \end{algorithmic}
\end{algorithm}

\smallskip
\noindent\textbf{Model Similarity Metric.}
In \sysname, we adopt the central kernel alignment (CKA)~\cite{kornblith2019similarity} to measure the similarity of any two models.
CKA is used for its better consistency in assigning similarity values to feature representations compared to other metrics such as cosine similarity \cite{kornblith2019similarity,jung2023feature}.
In addition, we will empirically evaluate the effectiveness of CKA through ablation studies.
The CKA similarity lies in the range $[0, 1]$, where smaller values indicate higher dissimilarity and larger values indicate higher similarity.
Taking $\mathrm{CKA}(\mathbf{X},\mathbf{Y})$ as an example, the CKA similarity is calculated as:
\begin{equation}
    \mathrm{CKA}(\mathbf{X},\mathbf{Y})=\frac{\mathrm{HSIC}(\mathbf{K},\mathbf{L})}{\sqrt{\mathrm{HSIC}(\mathbf{K},\mathbf{K})\mathrm{HSIC}(\mathbf{L},\mathbf{L})}},
\end{equation}
where $\mathbf{K=XX^\top}$, $\mathbf{L=YY^\top}$, and $\mathrm{HSIC}(\cdot, \cdot)$ denotes the Hilbert-Schmidt Independence Criterion (HSIC) value of two distributions \cite{gretton2005measuring}.
Further, the HSIC value is calculated as:
\begin{equation}
    \mathrm{HSIC}(\mathbf{K},\mathbf{L}) = \frac{1}{(n-1)^2}\mathrm{tr}(\mathbf{KHLH}),
\end{equation}
where $\mathrm{tr}(\cdot)$ is the trace of a matrix, $\mathbf{H} = \mathbf{I} - \frac{1}{n} \mathbf{1} \mathbf{1}^\top$ is the centering matrix, $\mathbf{I}$ is the identity matrix, and $\mathbf{1}$ is a vector of all ones \cite{gretton2007kernel}.

\subsection{Local Training and Global Aggregation}
\label{sec: local training and aggregation}

The procedure of client local training and server global aggregation is presented in Algorithm \ref{alg:alg1}.

\smallskip
\noindent\textbf{Overall Objective.}
The overall objective for the $i$-th client during the $t$-th round of FL is expressed as:
\begin{equation}
    \mathcal{L} = (1-\gamma)\mathcal{L}_a + \gamma \mathcal{L}_b + \mu \mathcal{L}_c,
    \label{eq: local overall loss}
\end{equation}
where $\gamma$ and $\mu$ are the hyper-parameters that determine the weights of the global adapter cross-entropy loss and the model-contrastive loss.
All the parameters in the backbone remain fixed throughout the training procedure.
In the $t$-th round, all trainable parameters are updated as follows:
\begin{equation}
    \label{eq: local update}
    (\theta_g, \theta_p, \phi_a, \phi_b) \gets (\theta_g, \theta_p, \phi_a, \phi_b) -\eta  \triangledown \mathcal{L}(\theta_g, \theta_p, \phi_a, \phi_b).
\end{equation}
For the $i$-th client, denoting $w_i = (\theta_g^i, \theta_p^i, \phi_a^i, \phi_b^i)$, the local objective is given by Eq.~(\ref{eq: local objective.}).

\smallskip
\noindent\textbf{FL Aggregation.}
For the FL aggregation, each client sends only the global adapter parameters to the server and retains the private adapter parameters locally.
The server computes a weighted average of the parameters received and broadcasts them to all clients for the next round of federated training.

\begin{table*}[t]
    \centering
    \caption{Performance of \sysname compared to all the baselines on cross-task, cross-silo datasets in the PEFT setting. The average and standard deviation of the accuracy (\%) are calculated over three runs. The \textbf{bold} and \underline{underline} fonts indicate the best and second-best results, respectively. The percentages of trainable parameters and communication overheads of each method w.r.t.~full FT are also presented.}
    \label{tab:overall}
    \begin{tabular}{l|ccccccc|cc}
        \hline
        Method & MRPC & RTE & SST-2 & QNLI & QQP & MNLI & Avg. & Param.~(\%) & Comm.~(\%) \\
        \hline
        FedAvg (Full FT) & 84.79$\pm$\scriptsize{1.29} & 77.46$\pm$\scriptsize{1.50} & 92.64$\pm$\scriptsize{0.50} & 88.40$\pm$\scriptsize{0.57} & 82.17$\pm$\scriptsize{1.23} & 73.94$\pm$\scriptsize{1.13} & 83.24$\pm$\scriptsize{0.22} & 100 & 100\\
        \hline
        Local & \underline{87.42}$\pm$\scriptsize{0.29} & 77.46$\pm$\scriptsize{0.83} & {93.63}$\pm$\scriptsize{1.77} & 87.09$\pm$\scriptsize{1.58} & 82.51$\pm$\scriptsize{1.50} & 73.37$\pm$\scriptsize{0.28} & 83.58$\pm$\scriptsize{0.22} & 1.16 & --\\
        \hline
        FedAvg (PEFT) & 87.09$\pm$\scriptsize{2.47} & 78.66$\pm$\scriptsize{1.81} & 93.30$\pm$\scriptsize{0.57} & 84.64$\pm$\scriptsize{1.98} & \underline{83.66}$\pm$\scriptsize{1.41} & 74.35$\pm$\scriptsize{1.58} & \underline{83.62}$\pm$\scriptsize{0.34} & 1.16 & 1.16 \\
        FedLR & 85.13$\pm$\scriptsize{1.02} & 74.58$\pm$\scriptsize{4.68} & 92.49$\pm$\scriptsize{1.02} & \underline{88.40}$\pm$\scriptsize{1.24} & 80.55$\pm$\scriptsize{3.68} & 72.39$\pm$\scriptsize{1.98} & 82.26$\pm$\scriptsize{0.95} & 0.58 & 0.58 \\
        FedAP & 86.60$\pm$\scriptsize{2.70} & 77.70$\pm$\scriptsize{1.25} & 93.47$\pm$\scriptsize{1.23} & 85.62$\pm$\scriptsize{1.23} & 81.37$\pm$\scriptsize{1.77} & 73.53$\pm$\scriptsize{2.14} & 83.05$\pm$\scriptsize{0.39} & 0.58 & 0.58\\
        MOON & 86.60$\pm$\scriptsize{0.75} & 78.90$\pm$\scriptsize{0.83} & 92.65$\pm$\scriptsize{1.77} & 85.62$\pm$\scriptsize{0.75} & 81.53$\pm$\scriptsize{1.23} & 73.20$\pm$\scriptsize{1.02} & 83.08$\pm$\scriptsize{0.86} & 1.16 & 1.16\\
        FedRep & 85.78$\pm$\scriptsize{0.49} & \textbf{79.14}$\pm$\scriptsize{1.90} & 92.65$\pm$\scriptsize{0.85} & 84.96$\pm$\scriptsize{2.32} & 81.37$\pm$\scriptsize{2.14} & \underline{75.82}$\pm$\scriptsize{1.98} & 83.29$\pm$\scriptsize{1.11} & 1.16 & 1.16 \\
        FedMatch & 87.09$\pm$\scriptsize{0.84} & 76.02$\pm$\scriptsize{0.81} & \underline{93.79}$\pm$\scriptsize{1.73} & 86.11$\pm$\scriptsize{1.26} & 83.33$\pm$\scriptsize{0.82} & 75.33$\pm$\scriptsize{1.25} & 83.61$\pm$\scriptsize{0.71} & 1.16 & 1.16 \\
        \hline
        \sysname (*Ours) & \textbf{87.42}$\pm$\scriptsize{0.83} & \underline{78.99}$\pm$\scriptsize{1.5} & \textbf{94.11}$\pm$\scriptsize{0.5} & \textbf{88.40}$\pm$\scriptsize{0.7} & \textbf{83.98}$\pm$\scriptsize{1.4} & \textbf{77.78}$\pm$\scriptsize{0.95} & \textbf{85.11}$\pm$\scriptsize{0.40}  & 1.16 & 0.58 \\
        \hline
    \end{tabular}
\end{table*}

\section{Experiments}

In this section, we conduct extensive experiments to evaluate the performance of \sysname.

\subsection{Dataset Construction}

In the experiments, we follow FedPETuning \cite{zhang-etal-2023-fedpetuning} to select six datasets, namely RTE, MRPC, SST-2, QNLI, QQP, and MNLI, from the GLUE benchmark \cite{radford2019glue}.
These datasets are widely used to evaluate the performance of natural language understanding (NLU) models, covering various tasks including textual entailment (RTE), sentiment classification (SST-2), sentence similarity judgment (MRPC and QQP), and semantic inference (QNLI and MNLI).

\smallskip
\noindent\textbf{Cross-task Cross-silo Setting.}
Our work is the first to establish a federated NLU dataset in a cross-task, cross-silo setting.
Unlike prior studies, we regard each of the six datasets as an independent client, ensuring data privacy during the training procedure.
To prevent larger datasets from dominating model training, we perform a random sampling on each dataset whose size is larger than MRPC to reduce its size to that of MRPC.

\smallskip
\noindent\textbf{Data Partitioning.}
As the GLUE benchmark does not release the test sets, we merge the existing training and validation sets, partitioning the dataset on each client into training, validation, and test sets in a 6:2:2 ratio. This dataset will be made publicly available to facilitate future work on federated NLU in cross-task, cross-silo settings.

\subsection{Baselines}
In the experiments, we compare \sysname with the following eight baselines:
\begin{itemize}
    \item \textbf{Local:} Each client trains a model locally without any communication with the server and other clients.
    \item \textbf{FedAvg \cite{pmlr-v54-mcmahan17a}:} The default FL method that trains a single global model for all participating clients. We use two variants of FedAvg: (\emph{i}) \emph{Full FT}, where all model parameters are updated and aggregated, and (\emph{ii}) \emph{PEFT}, where only the adapter parameters are updated and aggregated.
    \item \textbf{FedAP \& FedLR:} Two representative federated PEFT methods for PLMs proposed in \cite{zhang-etal-2023-fedpetuning} based on the adapters in \cite{houlsbyadapter2019} and \cite{hu2021lora}, respectively.
    \item \textbf{MOON \cite{li2021moon}:} A model-contrastive method that minimizes the distance between the representations learned by the local models and the global model.
    \item \textbf{FedRep \cite{collins2021FEDREP} \& FedMatch \cite{chen2021fedmatch}:} Typical personalized FL methods that capture shared and private knowledge.
\end{itemize}
To ensure a fair comparison in the PEFT setting, all baselines except FedAvg (Full FT) only fine-tune the additional adapter modules while keeping the architecture and parameters of the backbone model the same as \sysname.

\subsection{Hyperparameters and Implementation}

We searched for the learning rate $\eta$ from $\{10^{-3}, 5 \times 10^{-4},$ $10^{-4}, 5 \times 10^{-5}\}$ and set $\eta = 5 \times 10^{-4}$.
We adjusted the coefficients $\gamma$ and $\mu$ for the backbone and contrastive losses in the ranges $[0.1, 0.9]$ and $[0.01, 0.2]$, respectively, and decided $\gamma = 0.5$ and $\mu = 0.05$.
All the results reported are those under the default hyperparameter setting.
We used RoBERTa-Base\footnote{\url{https://huggingface.co/FacebookAI/roberta-base}} as the model backbone.
To accommodate the characteristics of various tasks, we did not share the classifier parameters across tasks.
The six clients participated in 25 communication rounds, training one epoch per round.
The bottleneck size of the adapters was set to 16.
We used the Adam optimizer with a batch size of 64.
All experiments were conducted on a Tesla V100 GPU with 32GB memory.

\subsection{Experimental Results}

\noindent\textbf{Overall Performance.}
Table~\ref{tab:overall} presents the performance of different methods in the federated cross-task, cross-silo setting.
First, \sysname achieves the best or second-best accuracy across all six clients.
This indicates that \sysname can adapt well to different characteristics of various data and tasks for NLU, exploiting both universal and specific knowledge to effectively personalize the model for each client.
Then, FedAvg (PEFT) performs better than FedAP and FedLR due to a greater number of trainable parameters.
However, FL methods without personalization (FedAvg, FedAP, and FedLR) are generally outperformed by local training in the cross-task, cross-silo setting, suggesting that personalization mitigates data and task heterogeneity issues in FL and provides most clients with better-performing models.
Finally, other personalized FL methods (FedRep and FedMatch) perform not significantly differently from FedAvg (PEFT).
This implies that they only handle data heterogeneity but do not consider task heterogeneity in the cross-silo setting.

\begin{figure}[t]
  \centering
  \includegraphics[width=0.48\textwidth]{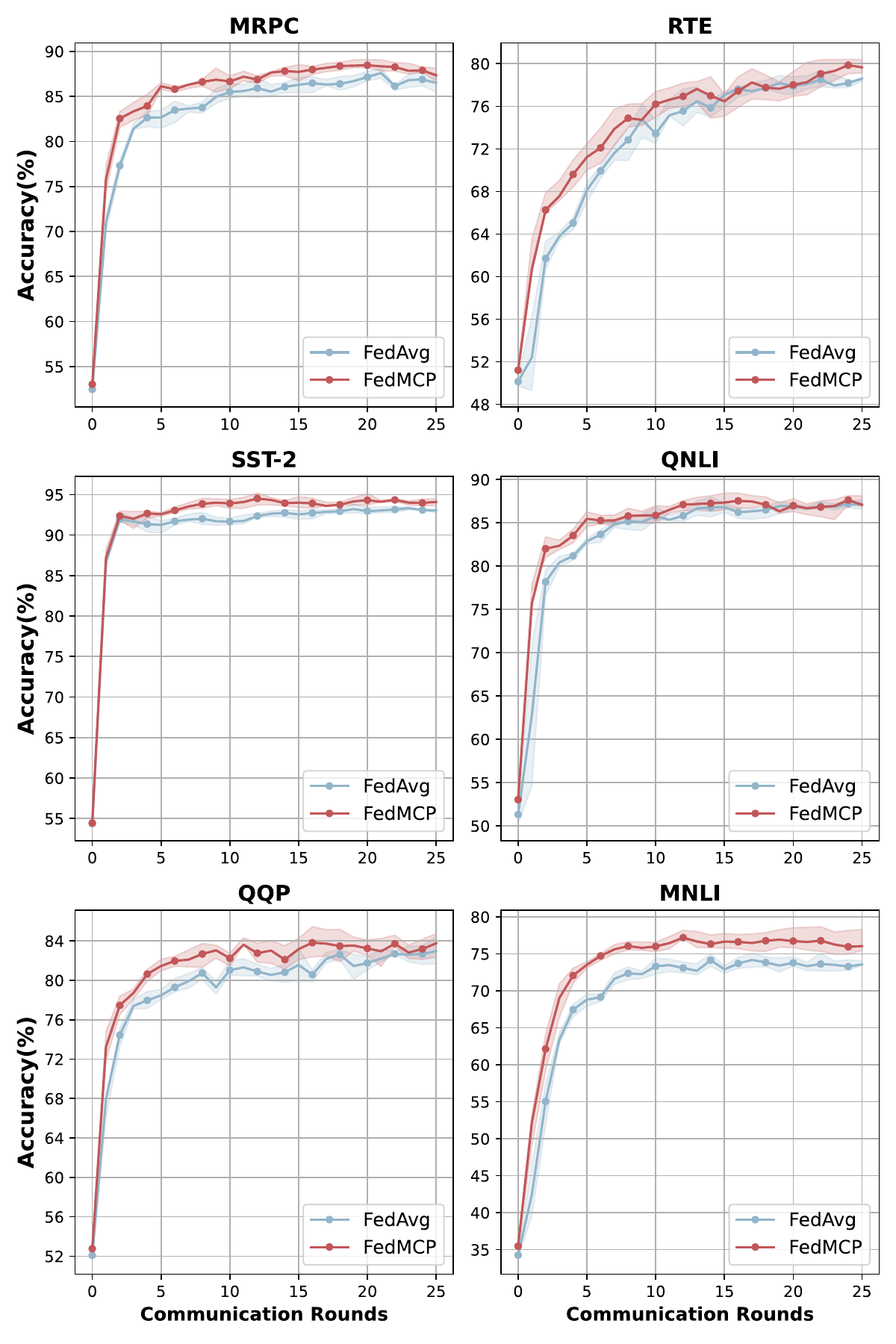}
  \caption{Comparison of \sysname and FedAvg (PEFT) for the average and standard deviation of accuracy during 25 communication rounds in six clients.}
  \label{fig:compare}
\end{figure}

In terms of efficiency, \sysname only updates 1.16\% of the model parameters and sends 0.58\% of them between the server and the clients in each communication round compared to FedAvg (Full FT) but still achieves better accuracy.
For all PEFT methods, the percentages of trainable parameters and communication overheads depend on the number of adapters added and used for aggregation (0.58\% for 1 and 1.16\% for 2).
The results confirm that \sysname strikes a better balance between parameter efficiency and accuracy than the baselines.
 
\smallskip
\noindent\textbf{Convergence Analysis.}
We performed a convergence analysis of \sysname in comparison to FedAvg (PEFT).
The average and standard deviation of accuracy during 25 communication rounds for each client are shown in Fig.~\ref{fig:compare}.
As both methods adopt the same model structure, the results reveal that \sysname achieves higher accuracy more rapidly than FedAvg (PEFT), with the same number of trainable parameters.
The faster convergence of \sysname suggests that the proposed model-contrastive learning and the structured loss function can effectively enhance personalized FL training.

\subsection{Ablation Studies}

In this subsection, we conduct ablation studies to investigate the effects of each component in the loss function, as well as the similarity metrics and sentence representations for model-contrastive personalization, on the performance of \sysname.

\smallskip
\noindent\textbf{Effect of Components in the Loss Function.}
As is shown in Eq.~(\ref{eq: local overall loss}), the two key components in the loss function of \sysname are the backbone loss (BL) and the contrastive loss (CL).
Table~\ref{tab: loss ablation} presents the average test accuracies for the six clients with three different loss functions: the entire one and those without BL and CL.
We observe that the average accuracy drops by 0.64\% when the BL is removed and 1.27\% when the CL is removed.
These results confirm the contributions of both components to \sysname: The BL can facilitate the learning of an effective global adapter to accommodate universal knowledge, and the CL can enable the private adapter to learn client-specific knowledge.

\begin{table}[t]
    \centering
    \caption{Effects of components in the loss function. Here, CL denotes the contrastive loss and BL denotes the backbone loss.}
    \label{tab: loss ablation}
    \begin{tabular}{c|ccc}
    \hline
    Method & FedMCP\textsubscript{w/o CL} & FedMCP\textsubscript{w/o BL} & FedMCP \\
    \hline
    Accuracy ($\%$) & 83.57$\pm$\scriptsize{0.68} & 84.13$\pm$\scriptsize{0.73} & \textbf{85.11}$\pm$\scriptsize{0.40} \\
    \hline
    \end{tabular}
\end{table}

\begin{figure}[t]
  \centering
  \includegraphics[width=0.36\textwidth]{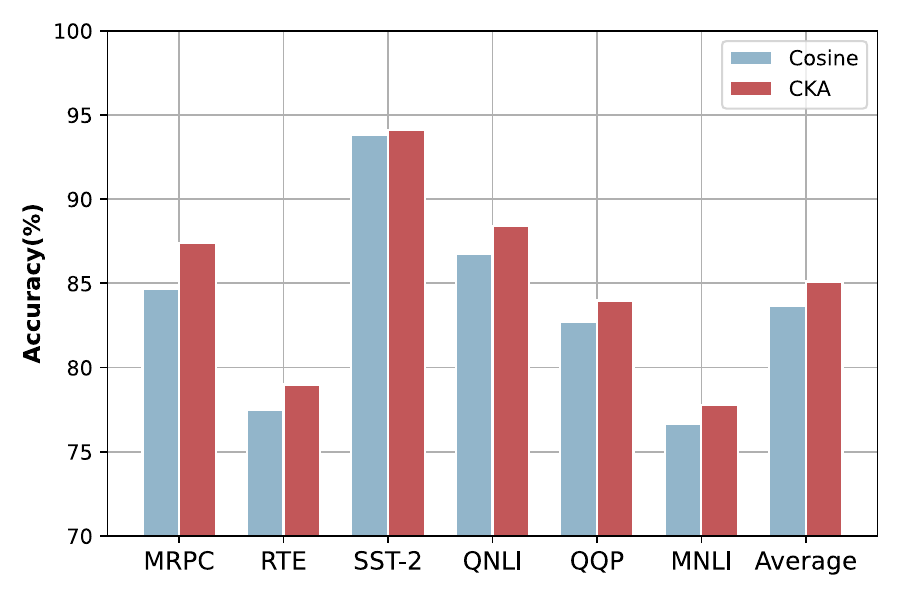}
  \caption{Effect of similarity metric (CKA vs.~cosine similarity) used in model-contrastive personalization on the performance of \sysname.}
  \label{fig: different_loss}
\end{figure}

\noindent\textbf{Effect of Similarity Metric in Model-Contrastive Personalization.}
We compare the performance of \sysname when CKA and cosine similarity are used as the similarity metric in model-contrastive personalization.
Fig.~\ref{fig: different_loss} illustrates the accuracy of the six clients using the two metrics.
The average accuracy when using cosine similarity is 83.66\%, which is 1.45\% lower than that when using CKA.
This decrease suggests that CKA is a more effective measure of model similarity in \sysname.
CKA might capture richer information than cosine similarity by assigning similarity values to feature structures.
Therefore, we use CKA in the implementation of \sysname.

\begin{figure}[t]
  \centering
  \includegraphics[width=0.36\textwidth]{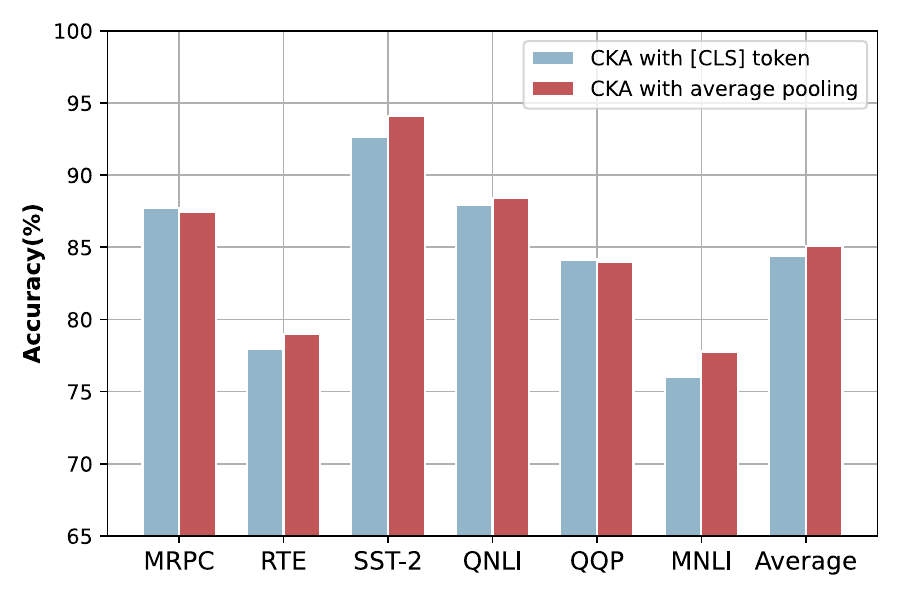}
  \caption{Effect of sentence representation ([CLS] token vs.~average pooling) used in model-contrastive personalization on the performance of \sysname.}
  \label{fig: different_token}
\end{figure}

\smallskip
\noindent\textbf{Effect of Sentence Representation in Model-Contrastive Personalization.}
For CKA similarity calculation, we use the average pooling of all tokens for sentence representation in \sysname.
An alternative approach is to use [CLS] tokens to represent the entire sentence, which is the default choice of BERT \cite{devlin2018bert} for sentence representation in text classification tasks.
Therefore, we investigate how both strategies affect the performance of \sysname.
The results are shown in Fig.~\ref{fig: different_token}.
The average accuracy with [CLS] token representations is 84.4\%, which is higher than the baselines in Table~\ref{tab:overall} but is 0.71\% lower than \sysname.
Although [CLS] tokens are designed to capture sentence semantics, they are optimized for classification tasks, potentially leading to more information loss in model-contrastive learning.
In contrast, the average pooling of all tokens provides more comprehensive sentence representations, which can better reflect the capacity of the model to learn sentence representations and distinguish between the global and client-specific knowledge for global and private adapters.

\section{Conclusion}

In this paper, we proposed a novel method, \sysname, for the PEFT of PLMs in cross-task, cross-silo FL.
\sysname could mitigate the non-IID issue and provide a personalized model specific to each client with distinct data and tasks using contrastive representations encoded in global and private adapters.
The model-contrastive method and the aggregation strategy of \sysname encouraged the global adapter to learn universal knowledge, reducing model drift between clients, and the private adapter to capture unique knowledge specific to each client.
Our experimental results showed that \sysname outperformed several baselines, including state-of-the-art personalized and PEFT FL methods for NLU tasks.

\bibliographystyle{IEEEtran}
\bibliography{mybibfile}

\end{document}